\documentclass[letterpaper, 10 pt, conference]{ieeeconf}  

\IEEEoverridecommandlockouts                              

\usepackage{graphicx} 
\makeatletter
\let\NAT@parse\undefined
\makeatother

\usepackage{hyperref}

\title{\LARGE \bf
Reasons People Want Explanations\\After Unrecoverable Pre-Handover Failures
}

\author{Zhao Han and Holly A. Yanco
\thanks{All authors are with the Department of Computer Science, University of Massachusetts Lowell, 1 University Ave, Lowell, MA 01854, USA
        {\tt\small zhan@cs.uml.edu, holly@cs.uml.edu}}%
}

\begin{document}

\maketitle
\thispagestyle{empty}
\pagestyle{empty}

\begin{abstract}
Most research on human-robot handovers focuses on the development of comfortable and efficient HRI; few have studied handover failures. If a failure occurs in the beginning of the interaction, it prevents the whole handover process and destroys trust. Here we analyze the underlying reasons why people want explanations in a handover scenario where a robot cannot pick up the object. Results suggest that participants set expectations on their request and that a robot should provide explanations rather than non-verbal cues after failing. Participants also expect that their handover request can be done by a robot, and, if not, would like to be able to fix the robot or change the request based on the provided explanations.
\end{abstract}

\section{INTRODUCTION}

Handing an object to someone else is seemingly easy if done by a human, but becomes challenging when a robot is the giver. The handover process can be separated into three phases \cite{strabala2013toward}: approach, signal, and transfer. The robot giver that possesses an object first approaches the human receiver, signals the intent that the robot is ready to hand over the object, and transfers the object to the receiver.

All three phases have been investigated for comfortable and efficient handovers, e.g., approaching a person behind a clustered table \cite{mainprice2012sharing} and in terms of arm trajectory and end effector height \cite{strabala2013toward}, object configuration \cite{cakmak2011human}, gaze effects on handover timing \cite{moon2014meet,zheng2015impacts}, and proactive release \cite{han2019effects}.

However, the assumption in all of this work is that the handover process is successful. There is work on robustness (e.g., robust transfer using force to detect object perturbation \cite{eguiluz2017towards,eguiluz2019reliable}), but unrecoverable handover failures remain unexamined. In this paper, we focus on the case where a robot has failed to pick up the object during the approach phase, preventing the whole handover process from happening.
In addition, early failures are shown to hurt humans' trust of the robot more than middle or later failures \cite{desai2013impact}.

To close the gap, we contribute a qualitative analysis on the textual data we collected during a human subjects study ($N=372$) about robot explanations. Particularly, we address the following question: \textit{Why and what would people want the robot to explain after a failure?}

\section{METHODOLOGY}

\subsection{User Study}

\begin{figure}[t]
	\centering
	\includegraphics[width=0.9\columnwidth]{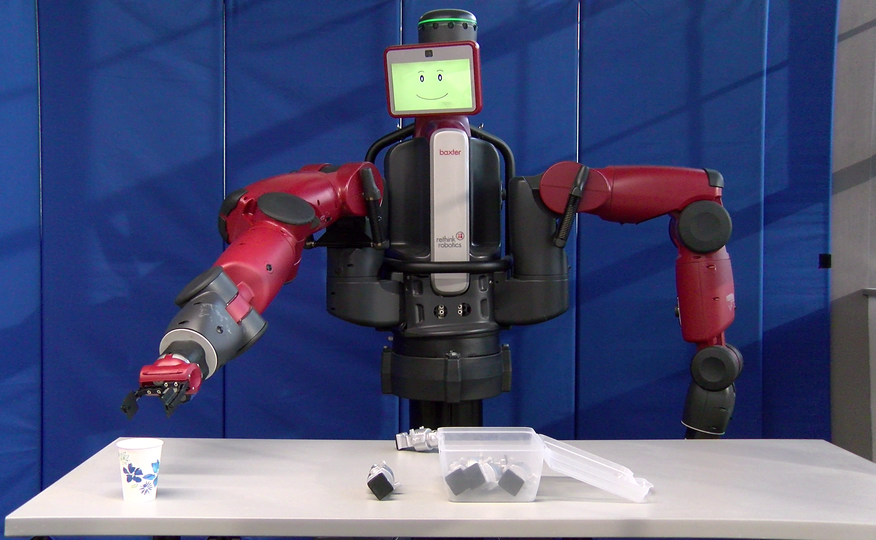} 
	\caption{The experiment scenario in which the robot was asked to pick up and hand over a cup that is slightly out of reach, a fact unknown to participants.}
	\label{fig:start} 
\end{figure}

We conducted an online experiment \cite{han2020need} on Amazon Mechanical Turk to investigate the perceived need and the content of desired robot explanations in a handover failure scenario. 372 participants contributed valid data (Age: 18--74, $M$ = 37; 210 males, 158 females, 3 who preferred not to answer, and 1 transgender person). In the experiment, participants first watched a Baxter robot encounter an unrecoverable pre-handover failure: it was unable to possess a cup when asked to hand it over. What participants were not told is the causal information that the cup slightly out of reach. 

We controlled the amount of causal information provided by how the robot executes the task and whether it shook its head (\textit{H}) or not (\textit{X}), resulting in 6 conditions. Executions included doing nothing (Clueless, \textit{C}), looking at the cup (Opaque, \textit{O}), and the addition of repeatedly moving its arm towards the cup (Legible, \textit{L}; see Fig. \ref{fig:start}). The latter two used non-verbal cues to hint or convey to participants that the cup is not reachable. All execution videos are available at \url{https://bit.ly/2U6VR0L}.

Surprisingly, participants report that the robot should explain regardless of the condition. Without explanations, the non-verbal cues are confusing to participants. The headshake was interpreted as disobeying whereas the intention of the arm movement was deemed unclear. For explanation content, the robot should explain why it failed, why it disobeyed them during head shake executions without any arm motion, and why it kept moving its arm, i.e., the intention. When the robot did nothing, people wanted to know about its previous behavior. A detailed accounting can be found in~\cite{han2020need}.

\subsection{Qualitative Analysis Approach}

Tightly related to this work, we also asked why participants would like the robot to explain. To analyze the qualitative data, we coded all open-ended responses. On a high level, we went through all the responses in two full iterations to develop the codes and revise them. Specifically, as we went through the responses, we first coded each with up to 4 codes and generalized them as we saw more responses. The generalization process led to frequent review of previous responses, especially during the first hundred. To account for later responses and limited working memory, we went through the responses again after the first iteration. After reading all of the responses, we revised the codes and created new composite codes that have an \textit{or} relationship with a very similar meaning.

\section{RESULTS \& ANALYSIS}

Of 372 participants, 353 answered the questions. After coding, we had 106 unique codes with half (55, 52\%) appearing only once due to the open-ended nature. We measured the inter-coder reliability using Cohen's $\kappa$. An independent coder coded 10\% of the samples, selected randomly, while the experimenter coded all responses. After merging codes with similar meanings, we achieved an $\kappa$ value of 0.84, considered as almost perfect agreement by \cite{landis1977measurement}.

\begin{figure}[t]
	\centering
	\includegraphics{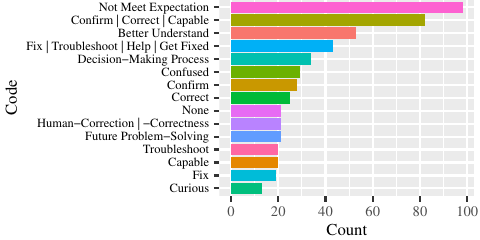} 
	\caption{Codes for explanation reasoning. Shown are those that appeared more than 12 times.}
	\label{fig:reason-bar} 
\end{figure}

\begin{figure}[t]
	\centering
	\includegraphics{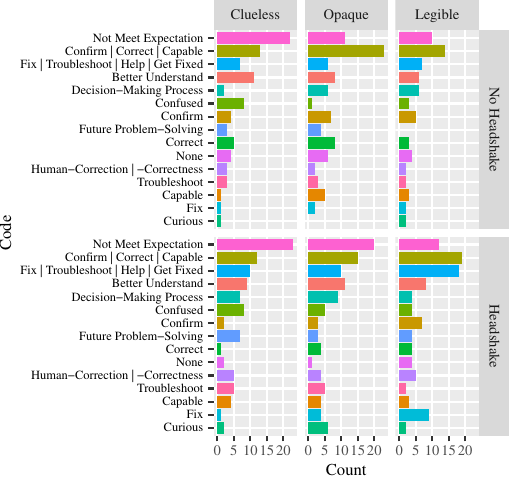} 
	\caption{Codes for explanation reasoning across conditions. Shown are those appeared more than 12 times.}
	\label{fig:reason-condition-bar} 
\end{figure}

Fig. \ref{fig:reason-bar} shows the top 15 codes that appeared more than 12 times for all 6 conditions, an average of 2 times per condition. Fig. \ref{fig:reason-condition-bar} shows the same data but across conditions.

Ninety-eight (26.3\%) participants wanted the robot to explain because the handover failure does \textit{not meet their expectation}. While there are 20 cases for Clueless conditions (CX, CH) and the Opaque condition with headshake (OH), there are only around 10 for the Opaque condition without headshake (OX) and Legible conditions (LX, LH). By comparing the Clueless and Opaque conditions, it shows that participants expect the robot to turn its head towards the cup (OX) but without a headshake (OH), which explains why the number in OX is dropped. The takeaway here is that \textit{participants set expectations of successful handovers from the robot after their request, and the robot should explain when it cannot achieve it}. By comparing the Opaque and Legible conditions, the additional arm movement does not lead to any change when there is no headshake, but the count reduces by half with a headshake. However, the reason why more participants want the robot to explain is to fix the robot or the second composite code, revealing how problems are perceived. The takeaway here is that \textit{when the robot cannot complete the task yet exhibits some unclear behaviors without explanation, participants interpret them as problems and that the robot needs to be fixed}. For other codes, the differences are not large across conditions, usually within 5--10, so we will not discuss them per condition below.

As seen in Fig. \ref{fig:reason-bar}, the second code by 82 participants (22\%) is a composite one: \textit{Confirm $|$ Correct $|$ Capable}, indicating that the robot should explain in order to confirm it will do the task, will do it correctly, and whether it is capable of finishing the task. Around 53 participants (14.3\%) expressed general reasoning: robot explanation helps them better understand the robot. Interestingly, in the fourth composite code: \textit{Fix $|$ Troubleshoot $|$ Help $|$ Get Fixed}, 43 participants (11.6\%) expressed interest in solving the problem of the robot, either by themselves or contact the manufacturer of the robot. Related to this, we found 21 participants (5.6\%) 
would like to correct themselves to make the robot work, coded as \textit{Human-Correction $|$ -Correctness}.

Due to the open-ended nature, all other codes found in the responses are fragmented and limited to less than 10\% of participants. Some interesting reasons include understanding the decision-making process and solving future problems. If they are given explicitly, more participants may choose them.

\section{CONCLUSIONS}

We explored the reasoning behind robot explanation when a robot cannot possess an object for a handover request. Results suggest that participants set expectations and the robot should not only use non-verbal cues but should explain after failing. Participants also showed interest in fixing the robot or correct themselves after getting robot explanations.

\section*{ACKNOWLEDGMENT}

This work has been supported in part by the Office of Naval Research (N00014-18-1-2503).

\addtolength{\textheight}{-17cm}   

\bibliographystyle{IEEEtran}
\bibliography{bib}

\end{document}